\documentclass{article}

\usepackage{arxiv}

\usepackage[utf8]{inputenc} 
\usepackage[T1]{fontenc}    
\usepackage{hyperref}       
\usepackage{url}            
\usepackage{booktabs}       
\usepackage{amsfonts}       
\usepackage{nicefrac}       
\usepackage{microtype}      
\usepackage{lipsum}		
\usepackage{multirow}
\usepackage{multicol}
\usepackage[justification=centering]{caption}
\usepackage{graphicx}
\usepackage{doi}
\usepackage[numbers]{natbib}

\title{License Plate Images Generation with Diffusion Models}

\date{} 					

\author{
	\href{https://orcid.org/0009-0000-3283-1659}{\includegraphics[scale=0.06]{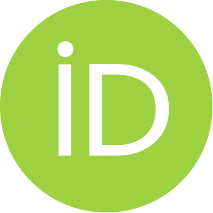}\hspace{1mm}Mariia Shpir} \\
    mariia.shpir@ukma.edu.ua\\ \\
	National University of Kyiv-Mohyla Academy, \\
	Kyiv, Ukrane. \\
    \And
	\href{https://orcid.org/0000-0001-8194-6196}{\includegraphics[scale=0.06]{orcid.pdf}\hspace{1mm}Nadiya Shvai} \\
    nadiya.shvai@cyclope.ai\\ \\
	Cyclope.ai, VINCI Autoroutes, \\
	Paris, France.\\ \\
	National University of Kyiv-Mohyla Academy, \\
	Kyiv, Ukrane. \\
	\And
	\href{https://orcid.org/0000-0001-9620-9324}{\includegraphics[scale=0.06]{orcid.pdf}\hspace{1mm}Amir Nakib} \\
    nakib@u-pec.fr\\ \\
	Cyclope.ai, VINCI Autoroutes, \\
	Paris, France.\\ \\
	University Paris Est Cr\'eteil, Laboratoire LISSI, \\
    Paris, France. \\
}

\renewcommand{\headeright}{}
\renewcommand{\undertitle}{}
\renewcommand{\shorttitle}{}

\hypersetup{
pdftitle={License plate images generation with diffusion models},
pdfsubject={Synthetic License Plate Generation using Diffusion Models for License Plate Recognition Tasks},
pdfauthor={Mariia Shpir, Nadiya Shvai, Amir Nakib},
pdfkeywords={License Plate Generation, Diffusion Models, Ukrainian License Plate Dataset},
}

\begin{document}
\maketitle

\begin{abstract}
	Despite the evident practical importance of license plate recognition (LPR), corresponding research is limited by the volume of publicly available datasets due to privacy regulations such as the General Data Protection Regulation (GDPR). 
To address this challenge, synthetic data generation has emerged as a promising approach. In this paper, we propose to synthesize realistic license plates (LPs) using diffusion models, inspired by recent advances in image and video generation. In our experiments a diffusion model was successfully trained on a Ukrainian LP dataset, and 1000 synthetic images were generated for detailed analysis.
Through manual classification and annotation of the generated images, we performed a thorough study of the model output, such as success rate, character distributions, and type of failures. Our contributions include experimental validation of the efficacy of diffusion models for LP synthesis, along with insights into the characteristics of the generated data. 
Furthermore, we have prepared a synthetic dataset consisting of 10,000 LP images, 
publicly available at \url{https://zenodo.org/doi/10.5281/zenodo.13342102}. 
%
Conducted experiments empirically confirm the usefulness of synthetic data for the LPR task. Despite the initial performance gap between the model trained with real and synthetic data, the expansion of the training data set with pseudolabeled synthetic data leads to an improvement in LPR accuracy by 3\% compared to baseline. 
\end{abstract}


\section{Introduction}
\label{sec:introduction} \

LPR plays important role in traffic management, automatic ticketing, security and surveillance. For the last decade, the state-of-the-art results in LPR have been achieved with deep learning methods \cite{laroca2022cross,khan2022automatic,padmasiri2022automated,wang2021rethinking,zou2022license}. 
The unprecedented success of neural networks (NN) in this and other computer vision tasks is based on the usage of large datasets. However, privacy protection laws such as GDPR \cite{regulation2016regulation} in European Union impose severe limitations on data collection, storage, and usage. Consequently, only a limited number of open LPR datasets are available, many of them including fewer than a thousand images \cite{weber_perona_2022, philip_updike_perona_2022, OpenALPR, MediaLab, UniversityOfZagreb2003}, particularly for those of European origin. 

One viable solution to restricted data availability is synthetic data. Template-based LP image generation \cite{silvano2021synthetic} is a robust and controllable approach that, nevertheless, requires meticulously handcrafted data augmentations to transform the template into a realistic image. Furthermore, it is highly dependable on the template itself and cannot be easily transferred to a similar problem. Generative adversarial networks (GANs) have allowed researchers to automate the template-to-real-image transformation \cite{wang2017adversarial, wu2019pixtextgan}, and to achieve higher diversity and real-life likeliness \cite{shvai2023multiple}. However, stability of GAN training remains a challenge.

Inspired by recent advances of diffusion models for image and video generation, 
we propose to apply them to LP synthesis.
%
%
To experimentally validate this approach, we trained a Denoising Diffusion Probabilistic Model (DDPM) \cite{ho2020denoising} on a Ukrainian LP dataset consisting of 78k images. Consequently, 1000 images were generated for a detailed analysis. They were manually classified as success and failure cases. Successfully generated LPs were annotated with the corresponding LP text. Based on these annotations, a character distribution analysis was performed. Furthermore, synthetic LPs were labeled for LPR task, and corresponding model was compared to the baseline trained with a real LP dataset.

Our contributions in this research can be summarized as follows:
\begin{itemize}
    \item we experimentally verify the viability of realistic LP synthesis with diffusion models;
    \item we analyze the outputs of the generative model, particularly the success rate, the character distribution, and the failure cases;
    \item we prepare a large synthetic dataset of 10,000 LP images that we release \cite{Synthetic_LP_Ukrainian} alongside this paper;
    \item we empirically confirm the usefulness of synthetic data for the LPR task.
\end{itemize}

The rest of the paper is organized as follows: Section \ref{sec:related_work} discusses the related work on LP generation and open LPR datasets. Section \ref{sec:methodology_results_discussion} provides details on the experiments methodology, results and their analysis. Section \ref{sec:dataset} briefly describes the dataset of synthetic Ukrainian LPs publicly released as a result of this work. Finally, Section \ref{sec:conclusions} concludes the article. Additional information regarding the dataset, including sample images, can be found in the Appendix.

\section{Related Work}
\label{sec:related_work}

\subsection{License Plate Generation}
\label{ssec:lp_generation} 
Generating synthetic LP images has become a practical solution to address the restrictions imposed by privacy protection laws on data collection and usage. Over the past few years, researchers have been investigating different methods to create realistic LP images for the purpose of training deep learning models in LPR tasks.

In the early days, computer graphics techniques were employed to generate LPs by creating artificial images using pre-designed templates and fonts. These methods are still being applied due to their robustness and controllability. Notably, Silvano et al. \cite{silvano2021synthetic} have proposed synthesizing Mersocur LP based on template and carefully designed image augmentations.  
These methods, although successful in producing significant amounts of data, frequently fell short in terms of diversity and realism required for training resilient LPR models. Moreover, they are difficult to extend due to their heavy dependence on a specific template.

Researchers have been investigating the potential of GANs in generating LP images that are more realistic. GAN-based approaches have demonstrated great potential in capturing the diverse range of factors present in real-world LPs, such as various fonts, lighting conditions, and backgrounds. 
In particular, Wang \textit{et al.} \cite{wang2017adversarial} have proposed an improvement of the CycleGAN based model \cite{zhu2017unpaired} that learns the mapping between the synthetic images generated by a script, and real images. 
Wu \textit{et al.} \cite{wu2019pixtextgan} have adopted a similar approach and improve the image-to-image based Pix2Pix generative architecture \cite{isola2017image}. 
Nevertheless, the challenges of generating high-quality images and ensuring a stable training process have been persistent in these approaches.

In recent times, diffusion models have gained significant attention as a compelling alternative for image generation tasks. Inspired by their success in producing high-quality photos and videos, we propose using diffusion models for LP synthesis. Diffusion models have numerous advantages, such as the capability to accurately model complex data distributions and produce a wide range of high-quality samples. Through this study, we explore the possibility of using diffusion models to create realistic Ukrainian LP images and examine their potential in tackling the challenge of limited data in LPR research.

\subsection{License Plate Datasets}
\label{ssec:lp_datasets}

Several LP datasets are openly available for research. They are commonly used for LP detection (LPD) or LPR tasks. We provide below their short description, and list them in Table \ref{tab:lp_datasets} with the information on number of images, their origin region, and type of the annotation available.
\begin{table*}[ht!]
\caption{Number of images, their origin and annotation type for some commonly used LP datasets; \\\textit{bbox} stands for \textit{bounding box}.}
\centering
\begin{tabular}{llll}
\hline
\textbf{Dataset name}                        & \textbf{\# of images} & \textbf{Origin} & \textbf{Annotations}                                                   \\ \hline
Caltech 1999   \cite{weber_perona_2022}          & 126    & USA     & None                        \\
Caltech 2001   \cite{philip_updike_perona_2022} & 526    & USA     & None                        \\
AOLP-AC  \cite{hsu2012application}                 & 681    & Taiwan  & None                        \\
AOLP-LE  \cite{hsu2012application}                 & 757    & Taiwan  & None                        \\
AOLP-RP \cite{hsu2012application}                  & 611    & Taiwan  & None                        \\
PKU-G1 \cite{yuan2016robust}                       & 810    & China   & LP bboxes           \\
PKU-G2 \cite{yuan2016robust}                       & 700    & China   & LP bboxes           \\
PKU-G3  \cite{yuan2016robust}                      & 743    & China   & LP bboxes           \\
PKU-G4  \cite{yuan2016robust}                      & 572    & China   & LP bboxes           \\
PKU-G5  \cite{yuan2016robust}                      & 1,152   & China   & LP bboxes           \\
OpenALPR-EU  \cite{OpenALPR}                       & 108    & EU      & \begin{tabular}[c]{@{}l@{}}LP bboxes, \\ LP text\end{tabular} \\
OpenALPR-US \cite{OpenALPR} & 222                       & USA             & \begin{tabular}[c]{@{}l@{}}LP bboxes, \\ LP text\end{tabular} \\
OpenALPR-BR \cite{OpenALPR} & 115                       & Brazil          & \begin{tabular}[c]{@{}l@{}}LP bboxes, \\ LP text\end{tabular} \\
SSIG \cite{gonccalves2016benchmark}                                                                & 2,000   & Brazil  &  LP characters bboxes     \\ 
CCPD  \cite{xu2018towards}                                                              &  $>$300,000 & China   &  \begin{tabular}[c]{@{}l@{}}LP bboxes, \\ LP text, \\ LP four vertices \end{tabular}                           \\
\begin{tabular}[c]{@{}l@{}}MediaLab LPR \\ database \cite{MediaLab} \end{tabular} 
& 181    & Greece  &     
\begin{tabular}[c]{@{}l@{}}None originally, \\ LP bboxes available in \cite{media-lab_dataset} \end{tabular}  \\ 
KarPlate Korea \cite{henry2020multinational}                                               & 3,893   & Korea   &   \begin{tabular}[c]{@{}l@{}}LP bboxes, \\ LP characters bboxes, \\ LP text\end{tabular}                          \\
University of Zagreb      \cite{UniversityOfZagreb2003}                                          & 509    & Croatia &   None                          \\
GAP-LP   \cite{kessentini2019two}                                                           & 9,175   & Tunisia  &   \begin{tabular}[c]{@{}l@{}}LP bboxes \\ LP characters bboxes \end{tabular}                          \\
UFPR-ALPR  \cite{laroca2018robust}                                                        & 4,500   & Brazil  &                              \begin{tabular}[c]{@{}l@{}}LP bboxes, \\ LP characters bboxes, \end{tabular}  \\
RodoSol-ALPR \cite{laroca2022cross}  & 20,000   & Brazil  &                             \begin{tabular}[c]{@{}l@{}}LP text, \\ LP layout, \\ LP four vertices \end{tabular} \\
IR-LPR-LPD  \cite{rahmani2022ir}  & 20,967 & Iran  &                             LP bbox  \\
IR-LPR-LPR  \cite{rahmani2022ir}  & 27,745 & Iran  &                             LP characters bboxes  \\
\hline
\end{tabular}
\label{tab:lp_datasets}
\end{table*}
Caltech Cars 1999 dataset \cite{weber_perona_2022} contains 126 images of cars from the rear. These images were taken in the Caltech parking lots. Caltech Cars 2001 \cite{philip_updike_perona_2022} is a car dataset containing 526 images of cars from the rear. They were taken on the freeways of southern California. 
AOLP dataset was composed by Hsu \textit{et al.} in \cite{hsu2012application} for LPD and LPR tasks. This dataset is divided into three distinctive subsets, namely access control (AC), law enforcement (LE), and road patrol (RP). AOLP-AC contains 681 images, AOLP-LE consists of 757 images, and AOLP-RP has 611 images. The origin of the data is Taiwan.
The Peking University LP dataset (PKU) contains in total 3,977 images split into five groups: G1, G2, G3, G4, G5 \cite{yuan2016robust}. Bounding boxes annotations for LP are also available, thus making it well suitable for LPD task.
OpenALPR \cite{OpenALPR} dataset has three subsets according to the data origin: Europian Union (OpenALPR-EU), USA (OpenALPR-US) and Brazil (OpenALPR-BR). OpenALPR-EU contains 108 images, OpenALPR-US has 222 images, and OpenALPR-BR consists of 115 images. Annotations include LP bounding boxes, and LP texts, thus making OpenALPR dataset suitable for LPD and LPR tasks.
SSIG benchmark dataset \cite{gonccalves2016benchmark} was collected by Gonçalves \textit{et al.} It is composed of 2,000 Brazilian LPs.  The corresponding 14,000 alphanumeric characters come with bounding box annotations.
CCPD proposed by Xu \textit{et al.} in \cite{xu2018towards} is a large dataset consisting of over 300,000 Chinese LP images. It has rich annotations that contain information on LP bounding box coordinates, LP text, LP four vertices locations, LP area \textit{wrt} the whole image, tilt degree, brightness and blurriness of the LP region.
MediaLab LPR database \cite{MediaLab}
consists of 181 Greek LP images. The dataset is split into multiple categories (day color images, day grayscale images, day images with blur, day images with shadows, day images with close view, and night capture). In addition, some difficult cases are presented, such as images with multiple vehicles, dirt and shadow, shadows in the LP, truck LP captured during the day, and truck LP captured during the night. Originally, no annotations were available for this data set; however, currently the corresponding LP bounding boxes can be found in the external source \cite{media-lab_dataset}.
University of Zagreb LP dataset was collected and released in course of License Plate Detection, Recognition and Automated Storage students' project in 2003 \cite{UniversityOfZagreb2003}. The image database has been prepared by using OLYMPUS CAMEDIA C-2040ZOOM digital camera. It contains over 500 images of the rear views of different vehicle types (cars, trucks, buses), taken under various lighting conditions (sunny, cloudy, rainy, twilight, night light). 
GAP-LP is a Tunisian LP database presented by Kessentini \textit{et al.} in  \cite{kessentini2019two}. The dataset consists of 9,175 images that were acquired with different quality cameras under different resolutions, view angles and daylight lighting conditions. The images are available at a dedicated website \cite{GAP-LP} together with LP bounding boxes and LP characters bounding boxes annotations. Consequently, GAP-LP is straightforward to be used for LPD and LPR tasks.
UFPR-ALPR is a Brazilian LP dataset presented by Laroca \textit{et al.} in  \cite{laroca2018robust}. It contains 4,500 images with LP bounding boxes and LP characters bounding boxes annotations, making it well suitable for LPD and LPR tasks. The images were obtained from three different cameras, 1,500 per each one. LPs in the UFPR-ALPR dataset belong to cars and motorcycles. 
The RodoSol-ALPR dataset contains 20,000 images from static cameras along the 67.5-kilometer ES-060 highway in Espírito Santo, Brazil, operated by Rodovia do Sol (RodoSol). Detailed in paper by Laroca \textit{et al.} \cite{laroca2022cross}, it features various vehicle types captured under diverse conditions, including day and night, different weather, and varying distances from the camera. Every image has the following information available in a text file: the vehicle type (car or motorcycle), the LP layout (Brazilian or Mercosul), LP text, LP four vertices coordinates (x,y).
IR-LPR is an Iranian LP dataset collected by Rahmani \textit{et al.} \cite{rahmani2022ir}. It includes 20,967 car images with the LP bounding box annotations. The total number of LP images with LP characters bounding boxes is 27,745 images.

Although some of the datasets detailed above are quite large, \textit{e.g.} CCPD containing over 300,000 images, or RodoSol-ALPR containing 20,000 images, European datasets are small and few: MediaLab LPR database of Greek LPs has only 181 images, OpenALRP-EU contains 108 images, and the University of Zagreb dataset has 509 images. Such scarcity is explained by the strict legislation on data privacy. To meet this challenge, we propose to generate synthetic LP images with DDPMs. We hope that the availability of a large number of realistic LP images will be beneficial for ANPR research.

\section{Experiments, Results and Discussion}
\label{sec:methodology_results_discussion}
In this section, we outline the experimental setup and conduct a thorough analysis of the generated images.
\subsection{Experimental Setup}
\label{ssec:experimental_setup} 

\subsubsection{Dataset}
\label{sssec:dataset}
\paragraph{Standardization of Ukrainian Vehicle License Plate Codes.} In 1995, Ukraine standardized its vehicle LP codes to include only 12 Ukrainian Cyrillic letters with Latin alphabet visual equivalents: A, B, E, I, K, M, H, O, P, C, T, X. Additionally, a two-digit regional code was added to the LP.

In 2004, numerical regional codes were replaced with new letter codes. In 2013, further modifications were made, replacing the initial letters in existing codes as follows: A to K (except Crimea, where AK became MA), B to H, and C to I. Table \ref{tab:license_plate_codes} presents the list of LP prefixes introduced in 2004, along with their corresponding 2013 codes and associated regions.

Since 2015, a new design featuring a blue band on the left side of the plate with the letters "UA" in white below the Ukrainian national flag was introduced. In 2021, Ukrainian regions were assigned two additional letter codes, although these have not yet been widely adopted. Furthermore, since 2020, vehicles powered by electric motors without internal combustion engines are designated with the characters Y or Z in their LP numbers.

Currently, all LPs issued in Ukraine, including Soviet-era plates, remain valid and can still be legally used on the country's roads. For this study, we will focus on one-line optimized plates of Type 1 (Regular vehicles) and Subtype 1-1-1 (Electric motor-powered vehicles), with codes registered between 2004 and 2021 \cite{ukraine_license_plates}. These specific types of plates are the most commonly observed and used in Ukraine.

\begin{figure}[ht]
    \centering
    \includegraphics[width=0.5\columnwidth]{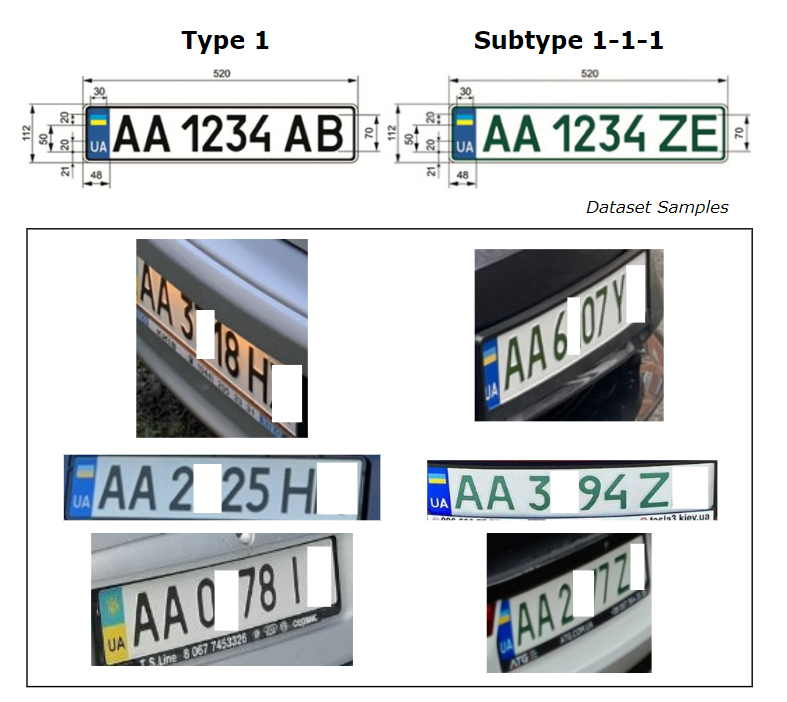}
    \caption{License plate formats and dataset samples. The top section shows the format types of Ukrainian LPs used in this study, and the bottom section provides corresponding sample images from the dataset.}
    \label{fig:license_plate_examples}
\end{figure}

\paragraph{Dataset Details.} Our study uses a private dataset of Ukrainian LP images, each captured in different lighting conditions and angles. In total, the dataset has 78,855 images (75,654 of Type 1, and  3,201 of Subtype 1-1-1), offering a wide range of real-world scenarios. See Figure \ref{fig:license_plate_examples} for more examples from the dataset.

\begin{table*}[hb!]
\caption{Correspondance of LP prefixes to regions of Ukraine.} 
\centering
\resizebox{0.4\textwidth}{!}{
\begin{tabular}{ccc}
\hline
\textbf{2004 Prefix} & \textbf{2013 Prefix} & \textbf{Region} \\
\hline
AA & KA & Kyiv city \\
AB & KB & Vinnytsia \\
AC & KC & Volyn \\
AE & KE & Dnipropetrovsk \\
AH & KH & Donetsk \\
AI & KI & Kyiv \\
AK & MA & Crimea \\
AM & KM & Zhytomyr \\
AO & KO & Zakarpattia \\
AP & KP & Zaporizhzhia \\
AT & KT & Ivano-Frankivsk \\
AX & KX & Kharkiv \\
BA & HA & Kirovohrad \\
BB & HB & Luhansk \\
BC & HC & Lviv \\
BE & HE & Mykolaiv \\
BH & HH & Odesa \\
BI & HI & Poltava \\
BK & HK & Rivne \\
BM & HM & Sumy \\
BO & HO & Ternopil \\
BT & HT & Kherson \\
BX & HX & Khmelnytskyi \\
CA & IA & Cherkasy \\
CB & IB & Chernihiv \\
CE & IE & Chernivtsi \\
CH & IH & Sevastopol city \\
\hline
\end{tabular}
}
\label{tab:license_plate_codes}
\end{table*}

\subsubsection{Generative Model Training}
\label{sssec:training} \

For our experiments, we used the Denoising Diffusion Probabilistic Model (DDPM), a class of generative models that iteratively reduce noise from a sample over a set number of steps. This process progressively transforms a sample from a noise distribution to a data distribution. We followed the training configuration proposed by the authors in the original DDPM paper, with specific adjustments tailored to our task.

Images were resized to 64x64 pixels without center cropping or flipping. For training, we used a batch size of 64. The chosen model consists of five feature map resolutions, ranging from 64x64 to 4x4, with dimensions [128, 128, 256, 256, 512]. The model includes two convolutional residual blocks per resolution level and self-attention blocks at the 16x16 and 8x8 resolutions between the convolutional layers.

The learning rate was initially set to $10^{-4}$, and subsequently changed using a cosine scheduler and a 5000-step warmup phase. Higher learning rates led to unstable training convergence. We used the AdamW optimizer with standard hyperparameters and applied EMA with a decay rate of 0.9999 to stabilize training. A dropout rate of 0.1 was applied to reduce overfitting.

The diffusion model was trained for a total of 100 epochs on a Tesla T4 GPU with 15GB of VRAM, which took approximately 30 hours to complete. For inference, the generated images were initially produced at a resolution of 64 × 64 pixels and later resized to 193 × 72 pixels to match the average aspect ratio of the training dataset. 

\begin{figure*}[b!]
    \centering
    \includegraphics[width=\textwidth]{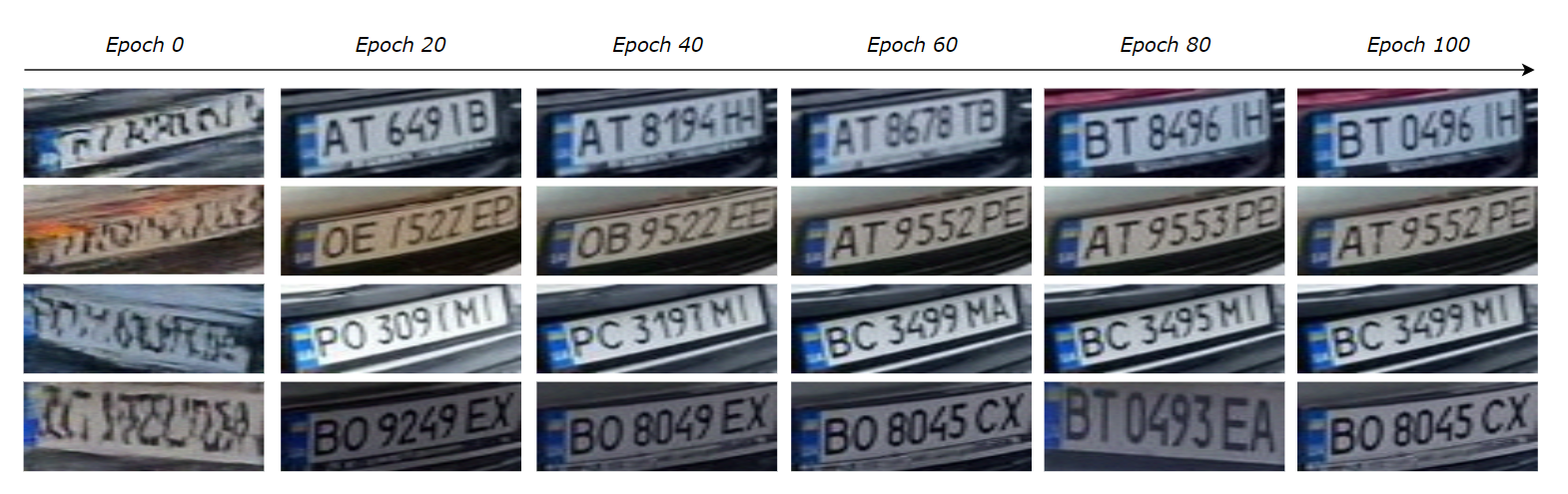}
    \caption{Visualization of the generated license plate images at different training stages.}
    \label{fig:generation_progress}
\end{figure*}

Figure \ref{fig:generation_progress} provides visualization of generated LP images over various training steps. We can clearly observe the progression: in the initial steps, the model learns the overall structure of the plates, and in the later steps, it adapts the text to match the target distribution.

\subsection{LP Generation Results}
\label{ssec:results} 

\paragraph{Successful Image Generation Criteria.}
For further analysis, we established criteria for \textit{successful image generation}. The criteria were based on the images being fully readable and meeting the following requirements:

\begin{itemize}
    \item The LP pattern must adhere to the format "AA0000AA," which is the standard LP format registered between 2004 and 2021.
    \item The prefix must correspond to a valid region code as per the Ukrainian regions listing (refer to Table \ref{tab:license_plate_codes}).
    \item The suffix is limited to the characters A, B, C, E, I, K, M, H, O, P, T, X, Y, and Z.
\end{itemize}

Images that do not meet these criteria were classified as \textit{failed image generation}.

\paragraph{Analysis of Image Generation Success and Failure Distribution.}
For the detailed analysis, we generated a total of 1,000 synthetic LP images using a trained diffusion model. Upon evaluation, 864 of these images were classified as successful (824 of Type 1 and 40 of Subtype 1-1-1), and 136 images were found to be unreadable or contained incorrect elements (see Figure \ref{fig:license_plate_examples_good} for successful examples and Figure \ref{fig:license_plate_examples_bad} for failed examples). 

\begin{figure}[h]
    \centering
    \includegraphics[width=0.6\columnwidth]{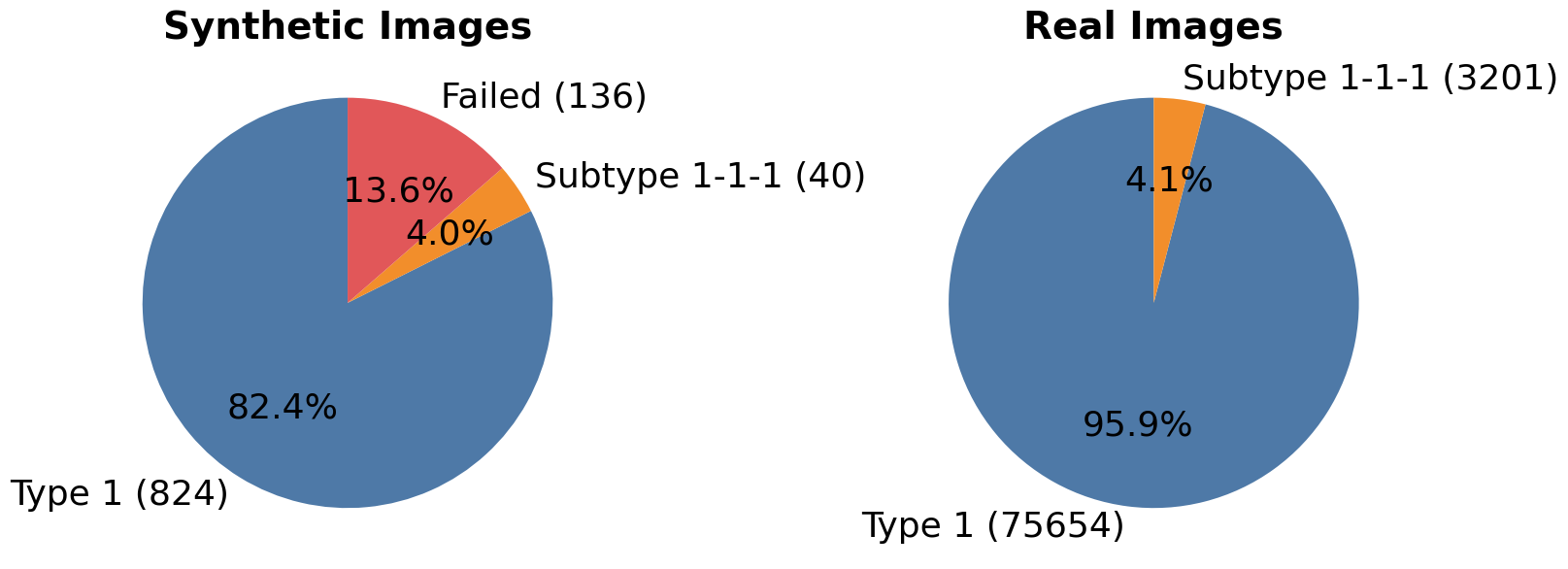}
    \caption{Distribution of image categories for synthetic (left) and real (right) LP images. }
    \label{fig:image_distribution}
\end{figure}

Additionally, Figure \ref{fig:image_distribution} provides a graphical representation of the distribution rates for Type 1, Subtype 1-1-1, and failure rates for both synthetic and real images in the LP generative model.

\begin{multicols}{2}
    \centering
    
    \includegraphics[width=1.015\linewidth]{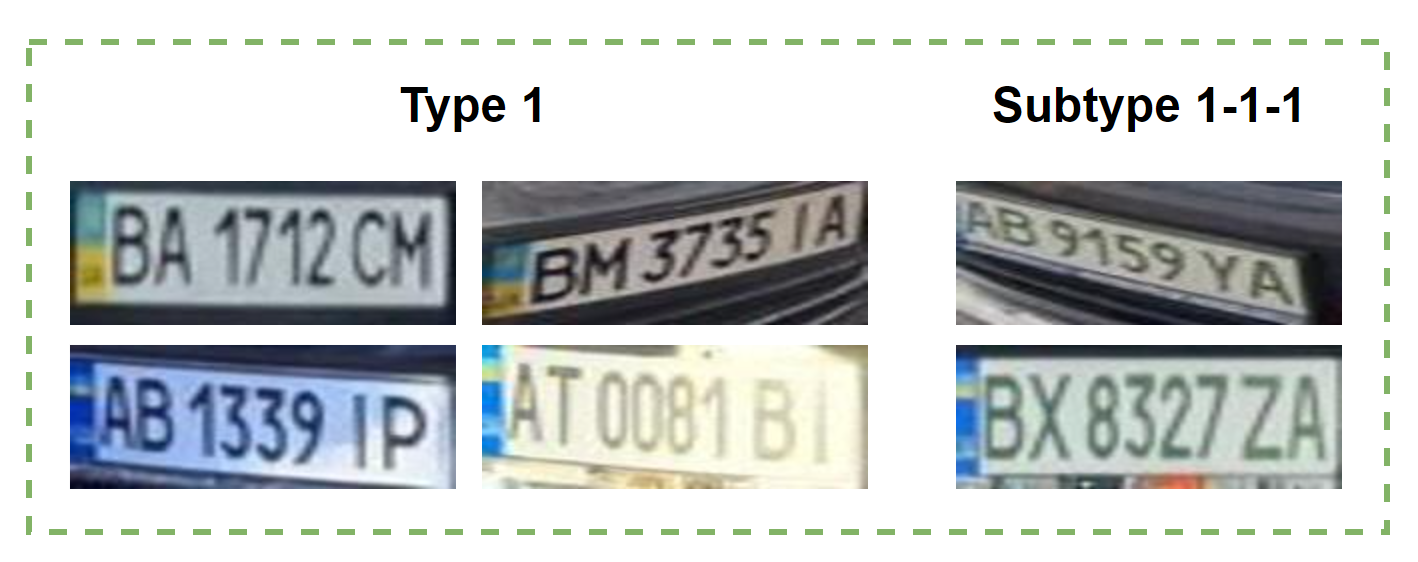}
    \captionof{figure}{Examples of successful LP image generation.}
    \label{fig:license_plate_examples_good}
    
    \includegraphics[width=1\linewidth]{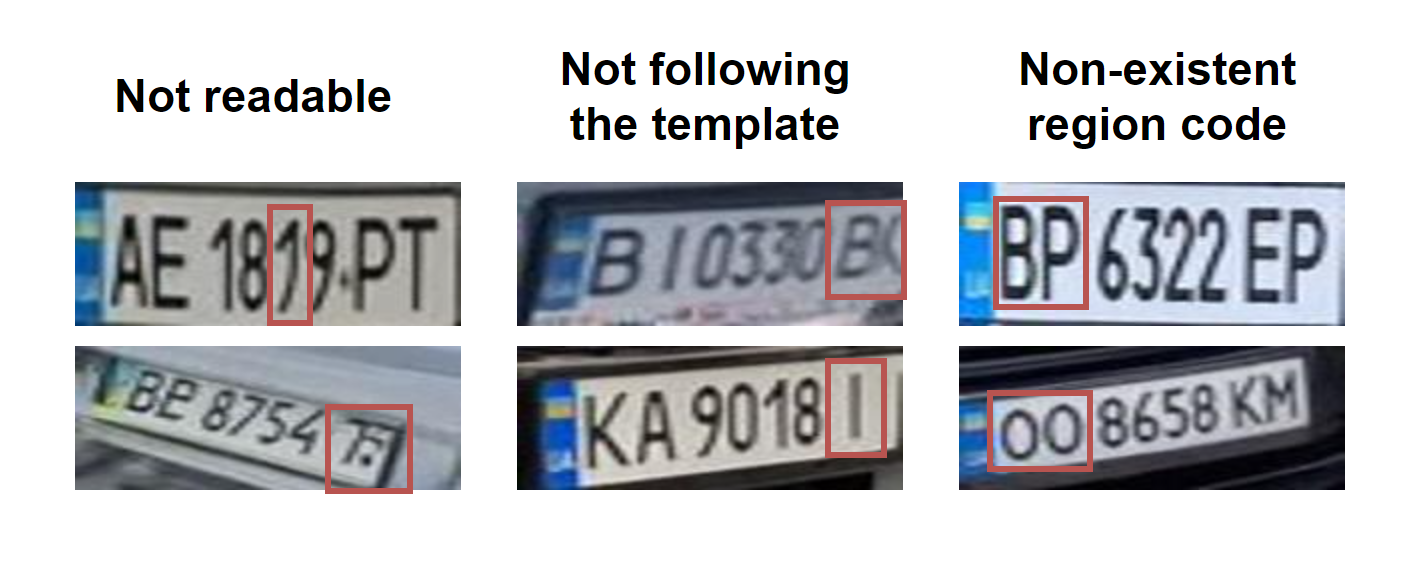}
    \captionof{figure}{Examples of failed LP image generation.}
    \label{fig:license_plate_examples_bad}
    
\end{multicols}

\paragraph{Quantitative Generation Quality Metrics.} To further assess the performance of LP generative model, we calculated the Fréchet Inception Distance (FID) \cite{heusel2018ganstrainedtimescaleupdate} scores for a generated synthetic dataset. The FID score is a widely used metric for evaluating the similarity between generated and real images. The FID score for a dataset of 15,864 synthetic images is 22.47.




\subsection{Character Distribution Analysis}
\label{sssec:char_distribution}

\begin{figure*}[tb]
    \centering
    \includegraphics[width=\textwidth]{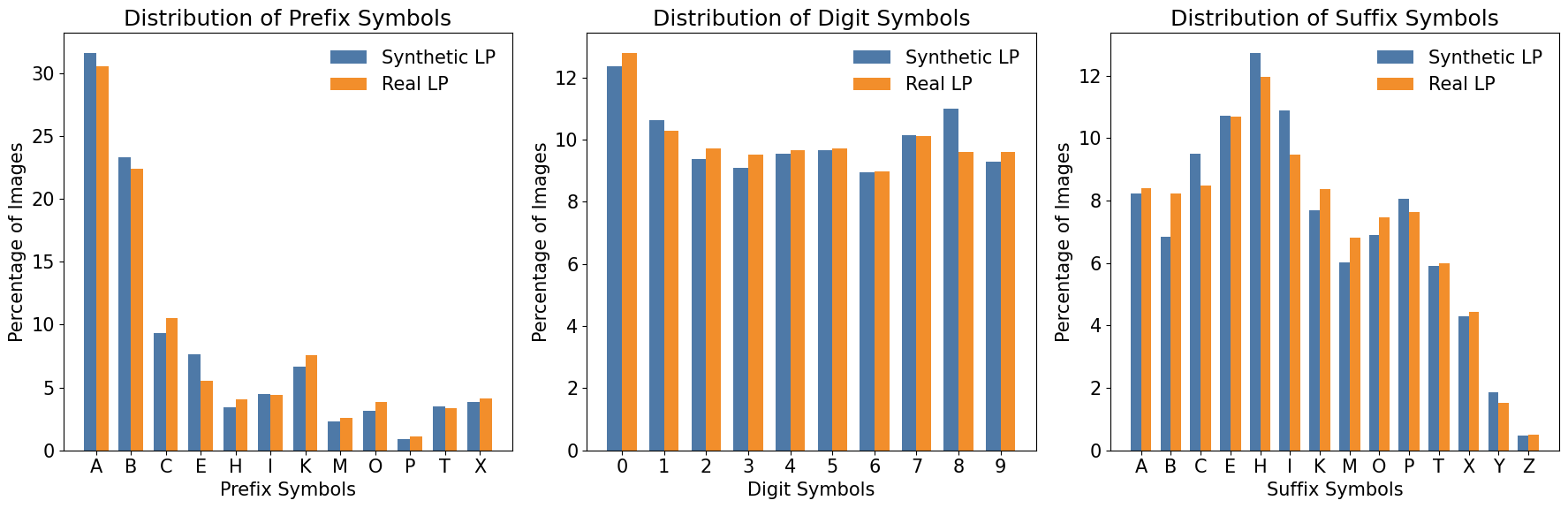}
    \caption{Comparison of prefix (left), digit (middle) and suffix (right) symbols' distributions in LP images.}
    \label{fig:symbol_distribution}
\end{figure*}

\paragraph{Symbol Distribution Analysis.} We analyzed the distribution of prefix, digit, and suffix symbols in synthetic and real LPs (see Figure \ref{fig:symbol_distribution}). The distributions show that synthetic data closely replicates real data, with only minor differences. For instance, the prefix symbol 'A' is highly frequent in both datasets, appearing in over 30\% of the images. The digit symbols are also similarly distributed, with '0' being the most common at around 12\% for both synthetic and real plates. However, some inconsistencies were observed.

Specifically, the digit '8' and the letter 'B' show differences in their frequencies. In synthetic LPs, '8' appears slightly more frequently than in real plates, and 'B' is more common as a suffix in generated LPs. This discrepancy could be due to the complexity in accurately drawing these symbols. The symbol 'I,' represented as a simple line, also appears more frequently in synthetic plates. This suggests that the model may prefer simpler shapes, which are easier to generate.

Interestingly, the variance in symbol distributions for prefix symbols is minimal compared to digits and suffix symbols. This is likely because there are fewer prefix combinations in the training data, leading to less variation in their synthetic representation. 

\begin{figure*}[t]
    \centering
    \includegraphics[width=\textwidth]{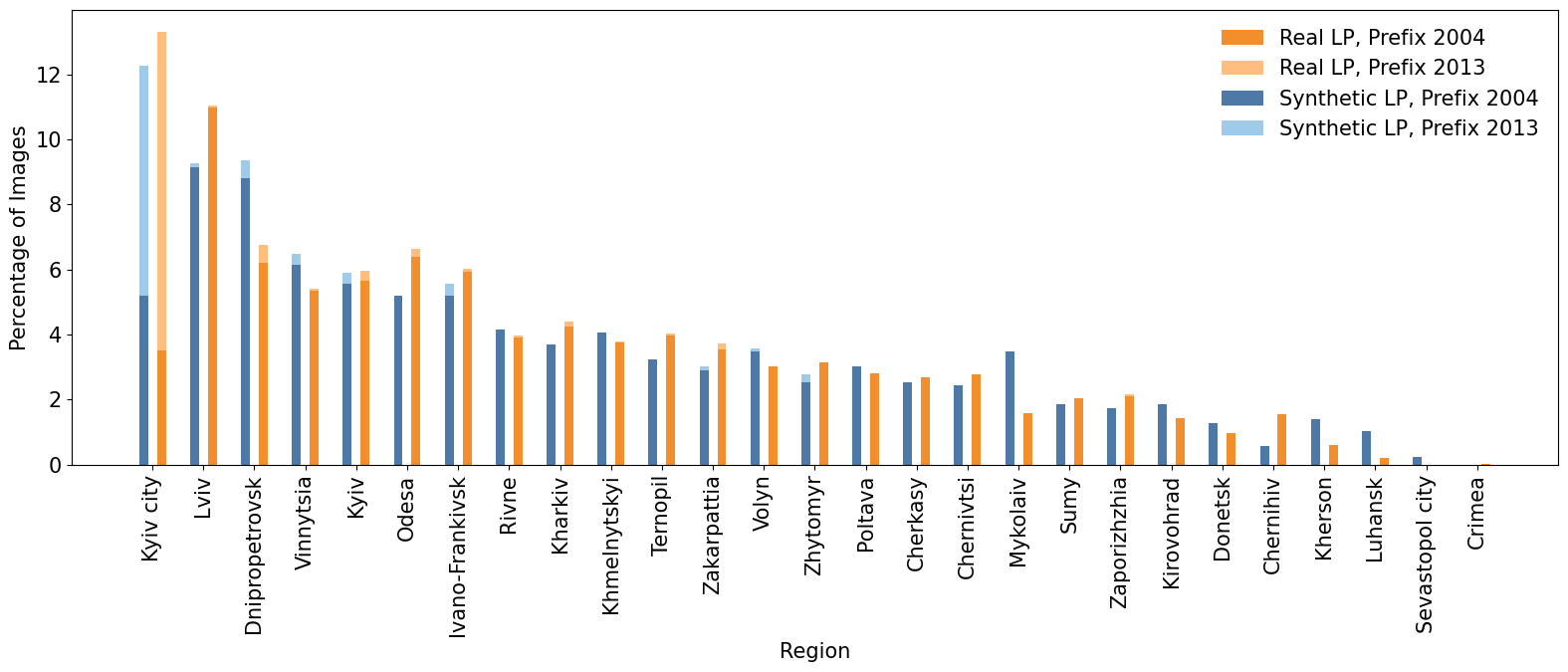}
    \caption{Comparison of region prefix distribution between synthetic and real datasets.}
    \label{fig:region_distribution}
\end{figure*}

\paragraph{Regional Distribution Analysis}
We also analyzed the regional distribution of LPs from different periods and types (synthetic and real) as shown in Figure \ref{fig:region_distribution}. Despite the complexity of regional data, the synthetic models show a high degree of similarity to the real data. However, some inconsistencies are present, which could be due to the model's difficulty in replicating the exact distribution of complex regional data. These variations highlight areas where the model could be refined for better accuracy.

Overall, despite these minor differences, the synthetic data generation process produces distributions that are largely similar to real-world data. 

\subsection{LPR with Synthetic Data}
\label{sssec:lpr}

\paragraph{Methodology for Training Character Detection Models.} To evaluate the usefulness of generated LP images for LPR task, we adopted the OCR-by-character-detection method prevalent in this domain \cite{rahmani2022ir, laroca2018robust, laroca2022cross}. 

We initiated the process by manually labeling 864 successfully generated synthetic LP images with bounding boxes. For a fair comparison, we also labeled 864 randomly sampled images from the training dataset. We then trained the YOLOv9-c model \cite{wang2024yolov9learningwantlearn}, chosen for character detection. The model trained on the real subset of images served as a baseline for accuracy comparison.

The YOLOv9-c model was trained using the Adam optimizer with a learning rate of 0.001 and a batch size of 16. We applied various image enhancement techniques, such as HSV adjustment, rotation, translation, scaling, shearing, and perspective transformation. To address the challenge of unbalanced character distribution, the training process incorporated image weights, allowing for image sampling proportional to the number of targets per image and inversely proportional to class frequency. The best model from this phase was selected based on the highest validation loss observed over 30 epochs. A dataset share of 10\% was reserved for validation.

For the synthetic data, we trained several instances of the model based on a gradual training dataset expansion with pseudolabeling. Initially, 864 synthetic images were manually annotated and used for training. Subsequently, an additional 5,000 synthetic images were pseudolabeled using the model trained on the initial 864 synthetic images, expanding the synthetic dataset to 5,864 images. In the final phase, 10,000 more synthetic images were pseudolabeled using the model trained on the 5,864-image dataset, bringing the total to 15,864 synthetic images.

To ensure high-quality image samples during pseudolabeling, we set the confidence threshold to 0.8. Each pseudolabeled image was selected based on the successful criteria described in Section \ref{ssec:results}. Each synthetic dataset was used to train separate models under the same setup as the baseline. It is important to note that pseudolabeling models are sensitive to overfitting, as overfitting can lead to the model ignoring failed parts of letters and focusing only on the main symbol features and their placement.

\paragraph{Validation and Testing of LPR Models.} 

Validation involved 200 real images to determine the optimal confidence threshold, maximizing LPR accuracy across the models. We considered the binary accuracy measure: 1 for a correctly read LP text and 0 for an incorrectly read LP text, regardless of the number of errors in the LP text or their type. The results of the validation experiments are shown in Table \ref{tab:validation_accuracy}. \label{tab:test_accuracy}

\begin{table*}[]
\caption{Validation results showing LPR accuracy across different confidence thresholds for models trained on the datasets. The bolded accuracy value represents the chosen optimal confidence threshold for each model, which was determined based on maximizing LPR accuracy.}
\centering
\renewcommand{\arraystretch}{1.2}
\begin{tabular}{|c|c|c|c|c|c|c|c|c|c|c|}
\hline
\multicolumn{2}{|c|}{\textbf{Training Dataset}} & \multicolumn{9}{c|}{\textbf{Accuracy at Confidence Threshold}} \\
\hline
\textbf{Size} & \textbf{Type} & 0.1 & 0.2 & 0.3 & 0.4 & 0.5 & 0.6 & 0.7 & 0.8 & 0.9 \\
\hline
864 & Real & 80.0\% & 89.5\% & \textbf{93.5\%} & 93.0\% & 91.0\% & 85.5\% & 79.0\% & 39.0\% & 0.0\% \\
864 & Synthetic & 72.5\% & 86.5\% & 92.0\% & 94.0\% & \textbf{94.5\%} & 89.5\% & 80.5\% & 53.8\% & 1.5\% \\
5864 & Synthetic & 84.0\% & 91.5\% & 94.5\% & 96.5\% & \textbf{96.5\%} & 95.5\% & 90.0\% & 71.5\% & 0.0\% \\
15864 & Synthetic & 95.5\% & 95.5\% & 97.5\% & \textbf{98.5\%} & 98.0\% & 97.0\% & 96.5\% & 91.5\% & 0.0\% \\
\hline
\end{tabular}
\label{tab:validation_accuracy}
\end{table*}

The test phase used 1,000 real LP images to evaluate the model performance. The models were applied with the optimal confidence thresholds found during the validation phase. The test results are given in Table \ref{tab:test_accuracy}. The baseline model, trained with 864 manually labeled real images, resulted in 94.1\% LPR accuracy. Its direct competitor, the model trained with 864 manually labeled synthetic images, showed a lower performance of 90.9\%. However, as the number of synthetic images increased, the LPR accuracy improved significantly. We observe that with equal manual labeling effort (of 864 images), generating and pseudolabeling more images significantly improves the LPR accuracy. The accuracy reaches 96.3\% with 5,864 synthetic images and further increases to 97.5\% with 15,864 synthetic images, both of which outperform the baseline accuracy of 94.1\% achieved by the model trained on real images. \

\begin{table*}[]
\caption{Performance comparison of LPR models on a test set of 1,000 real LP images. The bolded accuracy indicates the highest performance achieved among the models.}
\centering
\renewcommand{\arraystretch}{1.2}
\begin{tabular}{|c|c|c|}
\hline
\multicolumn{2}{|c|}{\textbf{Training Dataset}} & \multirow{2}{*}{\textbf{Test Accuracy}} \\
\cline{0-1}
\textbf{Size} & \textbf{Type} & \\
\hline
864 & Real & 94.1\% \\
864 & Synthetic & 90.9\% \\
5864 & Synthetic & 96.3\% \\
15864 & Synthetic & \textbf{97.5\%} \\
\hline
\end{tabular}
\label{tab:test_accuracy}
\end{table*}

These results demonstrate the high potential of synthetic data for enhancing LPR accuracy in scenarios with limited data availability, with a clear benefit from increasing the volume of synthetic training data.

\section{Dataset Description}
\label{sec:dataset}

To address the challenge of data scarcity in LPR tasks and to fill the gap in available Ukrainian LP datasets, we are releasing a dataset of 10,000 synthetic Ukrainian vehicle LP images \cite{Synthetic_LP_Ukrainian}. This dataset, which was instrumental in the second step of our dataset expansion process and contributed to training the highest-performing model in our experiments, is intended to serve as a robust resource for the training and validation of LPR models. By providing diverse conditions, including variations in lighting and angles, this dataset supports the exploration of model performance in real-world scenarios, ultimately aiding the development of more accurate and resilient LPR systems.

\paragraph{Dataset Composition}
The dataset contains 10,000 images, each with a resolution of 193 × 72 pixels. These images represent two main types of Ukrainian LPs: regular vehicles and electric motor-powered vehicles. The dataset covers a wide range of scenarios, including variations in lighting conditions, viewing angles, and regional codes, ensuring coverage of the standard LP formats used in Ukraine between 2004 and 2021. 

\textbf{Note:} As the data samples are synthetically generated, there may be slight inaccuracies in the representation of the intended distance between the letters and the exact color of the LPs. While these aspects have been approximated to closely resemble real-world conditions, they might not perfectly match the specifications of actual LPs.

\paragraph{Data Annotation}
Each image in the dataset is annotated in the YOLO Darknet format, which includes precise bounding box coordinates for each character on the LP. The annotations follow the standard LP format "AA0000AA," where:
\begin{itemize}
    \item \textbf{AA} represents the regional code.
    \item \textbf{0000} represents the numerical sequence.
    \item \textbf{AA} represents the suffix, corresponding to specific Ukrainian Cyrillic letters with Latin equivalents.
\end{itemize}

\paragraph{Distribution Analysis}
We conducted a detailed analysis of the character and regional distributions within the dataset. The frequency of digits and letters at each LP position is visualized in heatmaps (see Appendix, Figures \ref{fig:char_distribution_synthetic}), while the regional distribution based on LP prefixes is shown in a bar chart (see Appendix, Figure \ref{fig:region_distribution_synthetic}).
Additionally, we have included sample images in the Appendix (see Figure \ref{fig:sample_images}) that highlight different scenarios and conditions, providing a better sense of the dataset's quality and variety.

\section{Conclusions}
\label{sec:conclusions}
In this paper, we have proposed a novel approach for synthesizing realistic LPs using diffusion models, addressing the challenge of limited data availability in LPR tasks. Through experimental validation, we have demonstrated the efficacy of diffusion models in generating synthetic LP images that closely resemble real-world data and their usefulness for the LPR task. Our analysis of success rates, character distributions, and failure cases provides valuable insights into the capabilities and limitations of the proposed approach.

Furthermore, the creation of a synthetic dataset comprising 10,000 LP images adds to the existing resources available for LPR research, offering researchers and practitioners access to a diverse set of data for training and evaluation purposes. By releasing this dataset alongside the paper, we aim to contribute to the advancement of LPR technology and foster further research in this area.

Looking ahead, future research directions could explore enhancements to the proposed synthesis approach, such as incorporating additional sources of variability to generate more diverse LP images. Additionally, investigating the generalization capabilities of models trained on synthetic data to real-world scenarios would be a valuable avenue for further exploration.

Overall, our work highlights the potential of synthetic data generation using diffusion models in augmenting data availability for LPR tasks and lays the groundwork for future advancements in this field.

\appendix
\begin{figure*}[ht!]
    \centering
    \includegraphics[width=0.8\textwidth]{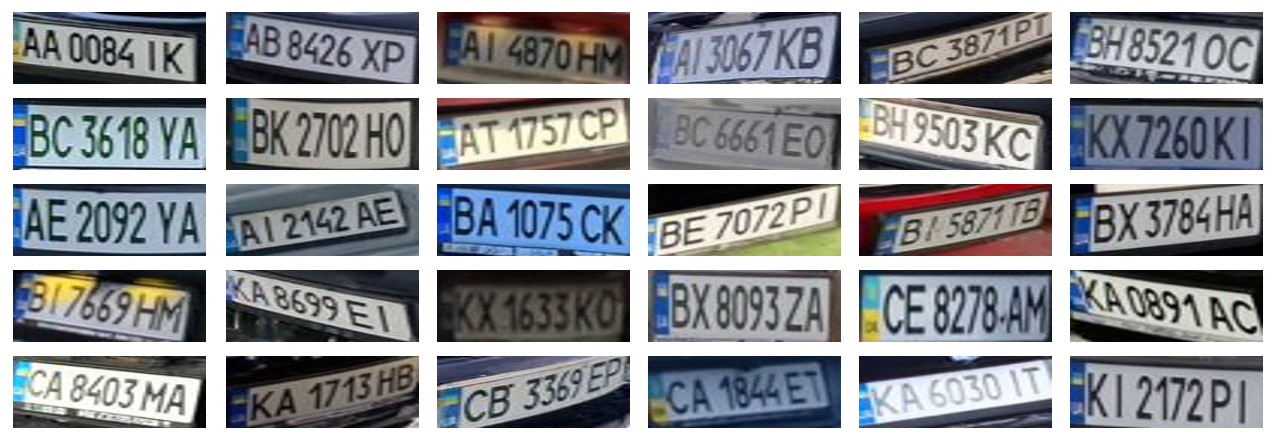}
    \caption{Sample images from the synthetic dataset. These images showcase the variety in lighting conditions, viewing angles, and regional codes present in the dataset.}
    \label{fig:sample_images}
\end{figure*}
\section{Distribution Analysis and Sample Images}

\subsection{Character Distribution by Position}
\begin{figure*}[h!]
    \centering
    \includegraphics[width=\textwidth]{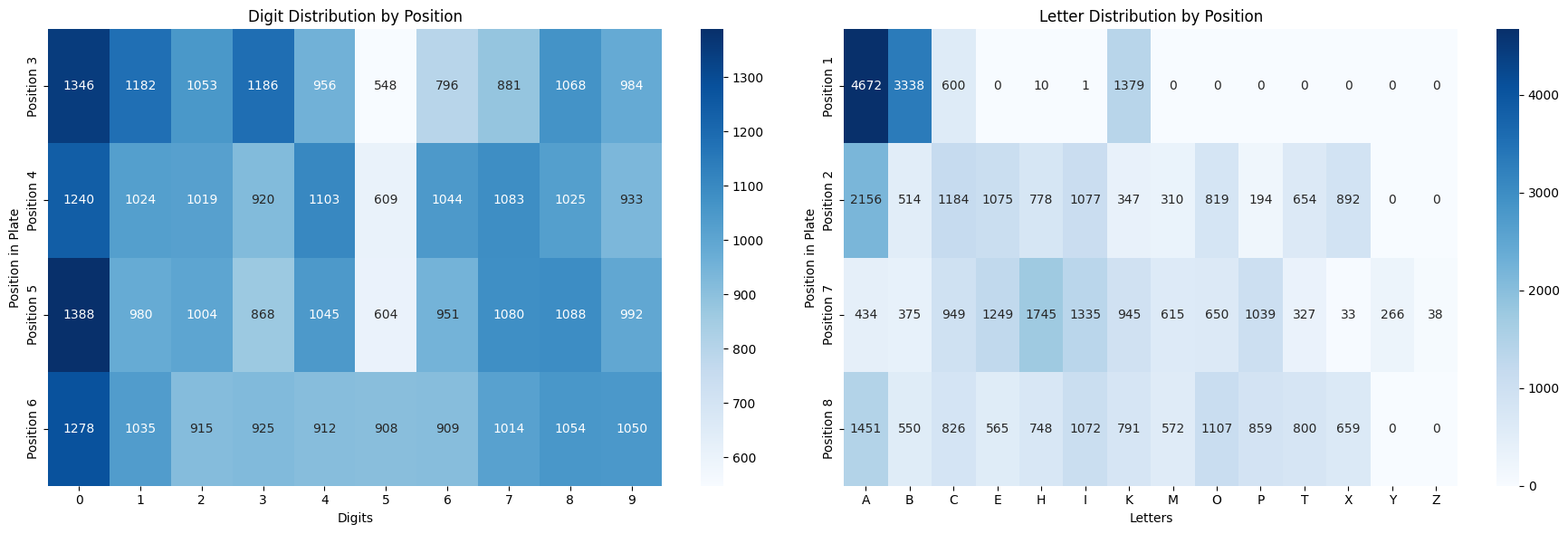}
    \caption{Character distribution by position in the provided Ukrainian LP synthetic dataset. The left heatmap displays the digit distribution, while the right heatmap shows the letter distribution across the license plates.}
    \label{fig:char_distribution_synthetic}
\end{figure*}

Figure \ref{fig:char_distribution_synthetic} shows the distribution of digits and letters across different positions in the synthetic license plates. The heatmaps highlight the frequency of each character at specific positions, revealing patterns that align with common license plate structures in Ukraine.

\subsection{Regional Distribution}
\begin{figure}[h!]
    \centering
    \includegraphics[width=0.8\textwidth]{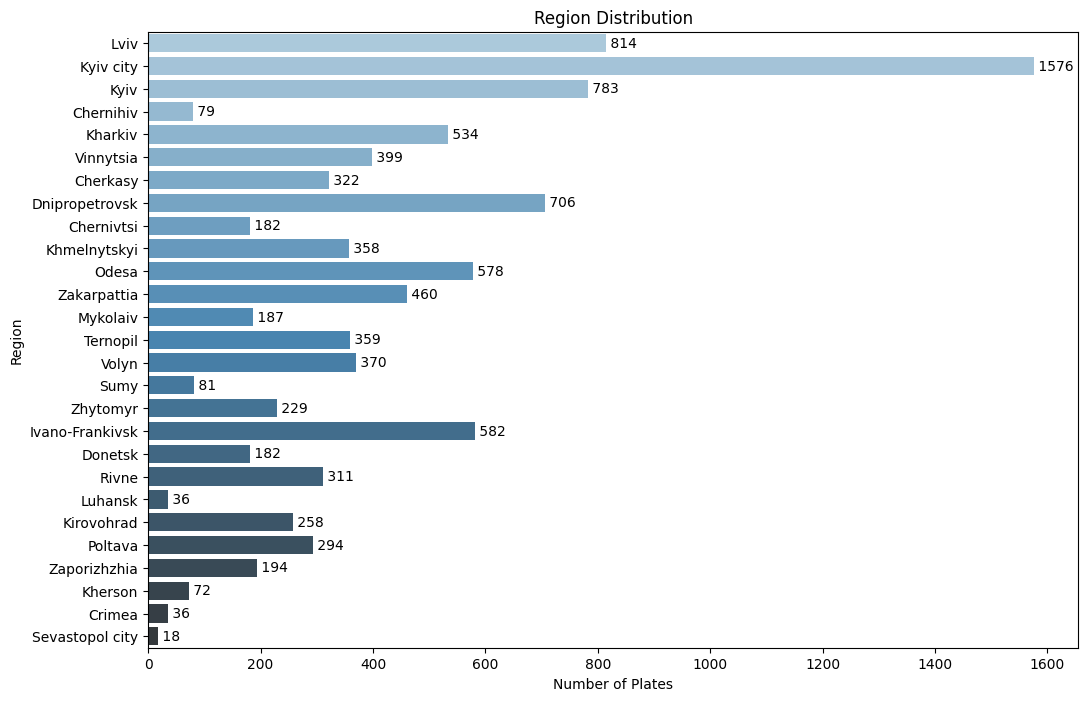}
    \caption{Regional distribution of License Plates in in the provided Ukrainian LP synthetic dataset. The bar chart shows the number of samples generated for each region, with the actual count displayed at the end of each bar.}
    \label{fig:region_distribution_synthetic}
\end{figure}

Figure \ref{fig:region_distribution_synthetic} presents the regional distribution of license plates based on their prefixes. This bar chart illustrates the number of license plates generated for each region, reflecting the dataset's geographic diversity.

\subsection{Sample Images}
To provide a clear understanding of the visual quality and diversity of the dataset, Figure \ref{fig:sample_images} includes a selection of synthetic license plate images. These samples represent various conditions such as different lighting, angles, and regional codes, demonstrating the dataset's capability to mimic real-world scenarios.

\bibliographystyle{unsrtnat}
\bibliography{main}

\end{document}